\documentclass[10pt,twocolumn,letterpaper]{article}

\usepackage{cvpr}
\usepackage{times}
\usepackage{epsfig}
\usepackage{graphicx}
\usepackage{amsmath}
\usepackage{amssymb}
\usepackage{multirow}
\usepackage{booktabs}
\usepackage{array}
\usepackage{authblk}
\usepackage{footnote}
\usepackage{bm}
\usepackage{bbm}
\usepackage{subcaption}

\usepackage[breaklinks=true,bookmarks=false]{hyperref}

\cvprfinalcopy 

\def\cvprPaperID{****} 

\begin{document}
\title{The Solution for CVPR2024 Foundational Few-Shot Object Detection Challenge}
\author{
Hongpeng Pan\textsuperscript{1},
Shifeng Yi\textsuperscript{1},
Shouwei Yang\textsuperscript{1},
Lei Qi\textsuperscript{1},
Bing Hu\textsuperscript{2},
Yi Xu\textsuperscript{2},
Yang Yang\textsuperscript{1,$\thanks{Corresponding Author}$}
}
\affil{
 $^1$Nanjing University of Science and Technology
 $^2$Dalian University of Technology
}

\maketitle
\begin{abstract}
This report introduces an enhanced method for the Foundational Few-Shot Object Detection (FSOD) task, leveraging the vision-language model (VLM) for object detection. However, on specific datasets, VLM may encounter the problem where the detected targets are misaligned with the target concepts of interest.
This misalignment hinders the zero-shot performance of VLM and the application of fine-tuning methods based on pseudo-labels. To address this issue, we propose the VLM+ framework, which integrates the multimodal large language model (MM-LLM). Specifically, we use MM-LLM to generate a series of referential expressions for each category. Based on the VLM predictions and the given annotations, we select the best referential expression for each category by matching the maximum IoU. Subsequently, we use these referential expressions to generate pseudo-labels for all images in the training set and then combine them with the original labeled data to fine-tune the VLM. Additionally, we employ iterative pseudo-label generation and optimization to further enhance the performance of the VLM. Our approach achieve 32.56 mAP in the final test.
\end{abstract}

\section{Introduction}

\begin{figure}
    \centering
    \includegraphics[width=0.85\columnwidth]{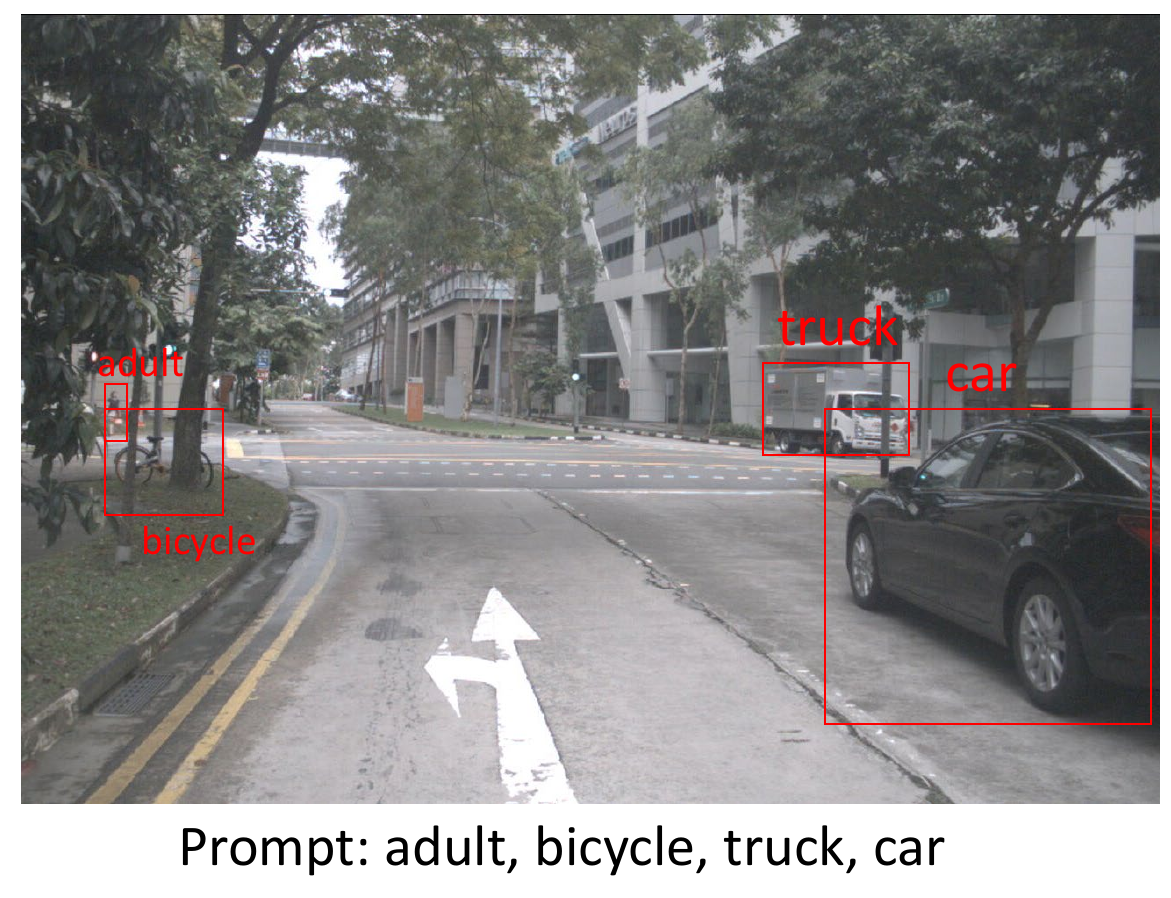}
    \vspace{-5pt}
    
    \caption{By specifying the interested classes in textual prompts, VLMs can implement zero-shot object detection.}
    \vspace{-10pt}
    \label{fig:intro1}
\end{figure}

Deep learning techniques have garnered widespread attention across multiple research fields.~\cite{YangHGXX23,YangWZL018, YangFZLJ21, YangWZX019, yang2021s2osc, xi2023robust, yang2024learning}. Object detection, as a fundamental task in computer vision, has garnered extensive research. Traditional visual recognition models are typically trained to predict a fixed set of predefined object categories, which limits their usability in real-world applications, as additional labeled data is required to generalize to new visual concepts and domains. To address this issue, some open-set object detection methods have been proposed, such as GLIP~\cite{LiZZYLZWYZHCG22} and Grounding DINO~\cite{abs-2303-05499}. These methods reframe object detection as a phrase-based task and introduce contrastive training between object regions and language phrases. Due to their excellent alignment between textual and visual features, these models are capable of performing object detection based on the provided prompts in a zero-shot manner, as illustrated in Figure~\ref{fig:intro1}.

\begin{figure}
    \centering
    \includegraphics[width=0.85\columnwidth]{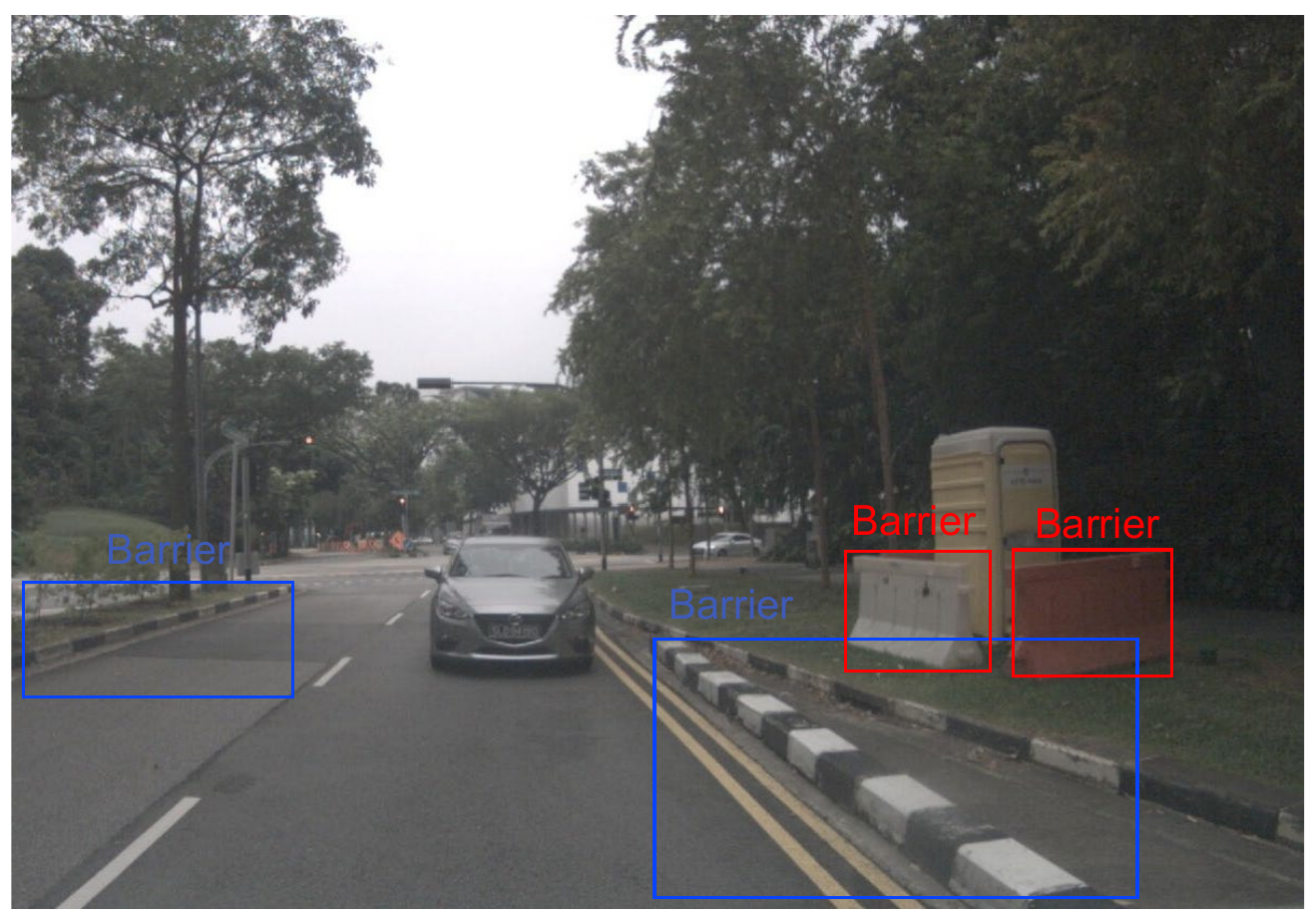}
    \vspace{-5pt}
    \caption{Poor Alignment Between VLM and Class Prompts. In the nuImages dataset, barriers are defined as road barricades (in \textcolor{red}{red}), while the obstacles predicted by the VLMs include roadside steps (in \textcolor{blue}{blue}).}
    \vspace{-10pt}
    \label{fig:intro2}
\end{figure}

\begin{figure*}
        \centering
    \includegraphics[scale=0.45]{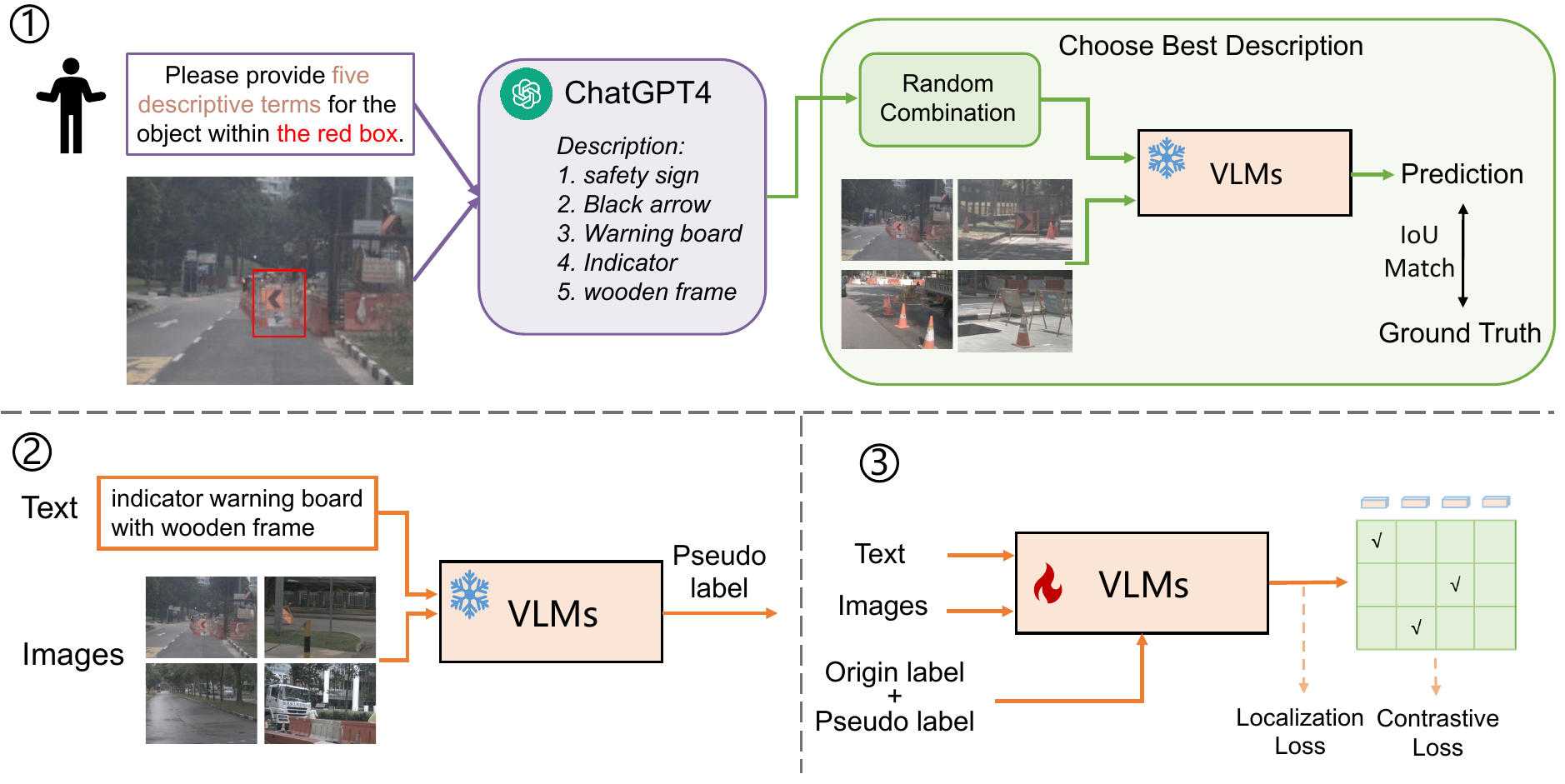}
    \vspace{-5pt}
    \caption{The framework of VLM+.}
    \label{fig: ourframework}
    \vspace{-10pt}
\end{figure*}
However, for specific target applications such as autonomous vehicle perception~\cite{CaesarBLVLXKPBB20}, these foundational models may still be suboptimal. This is primarily due to the challenge of aligning the foundational models with specific target concepts as shown in Figure~\ref{fig:intro2}. Under the Few-Shot Object Detection setting in~\cite{abs-2312-14494}, this misalignment can have a significant impact on methods that rely on predicting pseudo-labels for images.
Therefore, our goal is to reduce the gap in understanding between visual and textual concepts by the VLM, thereby minimizing the potential for generating erroneous pseudo-labels.
We propose a multi-stage approach named VLM+. Specifically, we first input images with annotations for each category into a MM-LLM~\cite{zhang2024mm}. Here, we use GPT-4~\cite{achiam2023gpt} to generate keyword prompts for these categories. Then, we randomly combine these prompts as referential expressions for the vision-language object detection model to obtain an optimal referential expression for each category, thereby enhancing the foundational model's understanding of the target concepts. Finally, we utilize the acquired referential expressions as textual input to improve the generation of pseudo-labels for each category. These pseudo-labels are then employed alongside the original labeled data as annotation data for the VLMs. Additionally, the trained model can be reused for pseudo-label generation and further optimization. The object detection VLMs we utilize comprise Grounding DINO and GLIP, with the corresponding pre-trained weights available at: \href{https://github.com/open-mmlab/mmdetection/tree/main}{https://github.com/open-mmlab/mmdetection/tree/main}.

\section{Method}
\subsection{VLMs}
\subsubsection{GLIP}
Open-set object detection is trained using existing bounding box annotations and aims to detect arbitrary classes through language generalization. GLIP~\cite{LiZZYLZWYZHCG22} considers the object detection task as a context-free phrase localization task, while phrase localization can be viewed as a context-aware object detection task. As a result, both can be improved within the same framework.
\subsubsection{Grounding DINO}

Grounding DINO~\cite{abs-2303-05499} is an open-set object detector that merges the Transformer-based detector DINO with grounded pre-training. This fusion enables the detection of arbitrary objects specified by human input, like category names or referring expressions. Grounding DINO lies in its feature fusion strategy across various stages of the detection pipeline. These include feature enhancers, text-guided query selection, and cross-modal decoders, effectively integrating textual and visual information.
\subsection{VLMs+}
\begin{table*}[htp]
    \centering
    \begin{tabular}{cllcc}
    \toprule
    Class index ($c$) & Class Name & Referential Expression ($T^c_{i^*}$) & ACC (before)& ACC (after)\\ \midrule
    1& car & car & 1.0&1.0\\ \hline
    2&truck & lorry & 0.6&  \textbf{0.7}\\ \hline
    3&construction vehicle & lift shovel excavator & 0.9&  0.9\\ \hline
    4&bus & bus & 0.9& 0.9\\\hline
    5&trailer & large cargo box on the trailer & 0.6& \textbf{0.8}\\\hline
    6&emergency & emergency police wagon & 0.4&\textbf{0.6}\\\hline
    7&motorcycle & narrow motorcycle & 0.8&  0.8\\\hline
    8&bicycle & bicycle bike & 0.9&  \textbf{1.0}\\\hline
    9&adult & adult people& 0.4&  \textbf{0.7}\\\hline
    10&child& single little short youth children & 0.6& \textbf{0.7}\\\hline
    11&police officer &  traffic policeman& 0.4&  \textbf{0.6}\\\hline
    12&construction worker& construction workman people& 0.5& \textbf{0.7}\\\hline
    13&personal mobility & small kick scooter & 0.3&  \textbf{0.9}\\\hline
    14&stroller & stroller & 1.0&  1.0\\\hline
    15&pushable pullable & pushable pullable garbage container & 0.5&  \textbf{1.0}\\\hline
    16&barrier & single short tarp barrier & 0.3&  \textbf{0.5}\\\hline
    17&traffic cone & traffic cone & 1.0&1.0\\\hline
    18&debris & indicator warning board with wooden frame  & 0.0& \textbf{0.7}\\\hline
    \toprule
    \end{tabular}
\vspace{-5pt}
\caption{Each class name, along with its corresponding referential expression. The VLM used is \href{https://github.com/open-mmlab/mmdetection/blob/main/configs/mm_grounding_dino/README.md}{Grounding DINO.} \textbf{Bold} indicates improved performance.}
\vspace{-5pt}

\label{tab:description}
\end{table*}
\subsubsection{Concept Alignment}
Regarding the misalignment between VLM and target concepts, we attribute it to the ambiguity and insufficient expression of category concepts. Therefore, to address this issue, we propose utilizing the image-to-text generation capabilities of a multimodal large language model to generate referential expressions that align with these concepts, instead of relying solely on class names, as depicted in Figure 3, labeled 1. Specifically, we overlay the annotation bounding box from the training set onto the corresponding images and set the prompt as: “Please provide five descriptive terms for the object within the red box.” as input of MM-LLM. Then, we can obtain a set of descriptive prompts for each category. Leveraging the language comprehension and vision-language alignment abilities of the vision-language model, we randomly combine five prompts into $N$ referential expressions. These expressions serve as inputs to the VLM, with the aim of selecting the best referential expression for the class name.
Below, we will discuss the process of selecting the optimal reference expression in detail.

Let \( T^c_i \) and \( P^c_{i,j} \) denote the \(i\)-th referential expression of the \(c\)-th class and the bounding box positions of the \(j\)-th image obtained after processing \( T^c_i \) through the VLMs, respectively. \( B^c_j \) represents the the \(j\)-th ground-truth bounding box of the \(c\)-th class, where \(c = 1, 2, \ldots, 18\), \(i = 1, 2, \ldots, N\), and \(j = 1, 2, \ldots, 10\).
To start, we compute the Intersection over Union (IoU) for each \( P^c_{i,j} \) with \( B^c_j \):
\[
\text{IoU}(P^c_{i,j}, B^c_j) = \frac{|P^c_{i,j} \cap B^c_j|}{|P^c_{i,j} \cup B^c_j|}
\]

To calculate the prediction accuracy of VLMs under the current referential expression, we first define an indicator function to check if the IoU is greater than 50\%:

\[
\mathbbm{1}_{\text{IoU}(P^c_{i,j}, B^c_j) > 0.5} =
\begin{cases}
1, & \text{if } \text{IoU}(P^c_{i,j}, B^c_j) > 0.5 \\
0, & \text{otherwise}
\end{cases}
\]

Next, we define the accuracy for each set of bounding boxes as:

\[
\text{ACC}(P^c_{i,j},B^c_j) = \frac{1}{10}\sum_{j=1}^{10} \mathbbm{1}_{\text{IoU}(P^c_{i,j}, B^c_j) > 0.5}
\]

We then find the set of bounding boxes with the highest accuracy:

\[
i^* = \arg\max_{i=1, \ldots, N} \text{ACC}(P^c_{i,j})
\]

Finally, we select the referential expression \( T^c_{i^*} \) as the best referential expression for $c$-th class name.
As shown in Table~\ref{tab:description}, we present the best referential expression obtained by the VLM for each class name. Additionally, we demonstrate the improvements in prediction accuracy of VLM for each category before and after using referential expressions. We can observe significant improvements in the recognition ability for classes such as ``personal mobility'', ``debris'', ``pushable pullable'' and ``trailer''. This significantly improves the model's predictive performance on categories within a specific dataset, and holds the promise of generating higher-quality pseudo-labels for subsequent model training, thereby achieving better fine-tuning results.

\subsubsection{Iterative Pseudo-label Optimization}
For the federated dataset provided by the competition, we implement the iterative pseudo-label optimization approach. Iterative pseudo-label optimization involve a process where pseudo-labels, predicted labels assigned to unlabeled data by the VLMs, are iteratively generated and refined. If the confidence score of the label generated by the model for a category exceeds pseudo-label threshold \(\eta\), we consider it as pseudo-labeled data for this category. Below, we outline the detailed process of iterative pseudo-label optimization.

\begin{enumerate}
\item \textbf{Initial Pseudo-Label Generation}: Initially, pseudo-labels are generated for unlabeled data using the initial model and referential expressions. These pseudo-labels serve as initial labels for the unlabeled data.

\item \textbf{Model Training}: The model is then trained on both labeled data with ground truth labels and pseudo-labels. This training process aims to improve the model's performance using the combined labels.

\item \textbf{Pseudo-Label Refinement}: After training, the model's predictions on unlabeled data are updated based on the new model parameters. These updated predictions serve as refined pseudo-labels for the next iteration.

\item \textbf{Iteration}: Steps 2 and 3 are repeated iteratively, with the model being retrained on the combined labels and the pseudo-labels being refined in each iteration. This iterative process continues until a convergence criterion is met or a predefined number of iterations is reached.

\end{enumerate}
\subsubsection{Loss Function}
The loss functions for Grounding DINO and GLIP include Focal Loss, box L1 loss, and GIOU loss. The weights for these losses are set to 1.0 for Focal Loss, 5.0 for box L1 loss, and 2.0 for GIOU loss. For Grounding DINO, similar to the DETR model, we add auxiliary losses after each decoder layer and encoder output.

\begin{table}[htp]
    \centering
    \begin{tabular}{l|l}
    \toprule
    Methods & mAP \\ \midrule
    Baseline (Best) &21.51 \\ \midrule
    GLIP (zero-shot) & 15.73 \\ \midrule
    GLIP + & 27.27\\ \midrule
    Grounding DINO (zero-shot) & 19.91\\ \midrule
    Grounding DINO + & 32.56\\ 
    \bottomrule
    \end{tabular}
\caption{Comparison of VLM and VLM+.}
\label{tab: compare}
\end{table}

\begin{figure}[htbp]
    \centering
    \begin{subfigure}[t]{1\linewidth}
        \centering
        \includegraphics[width=0.49\textwidth]{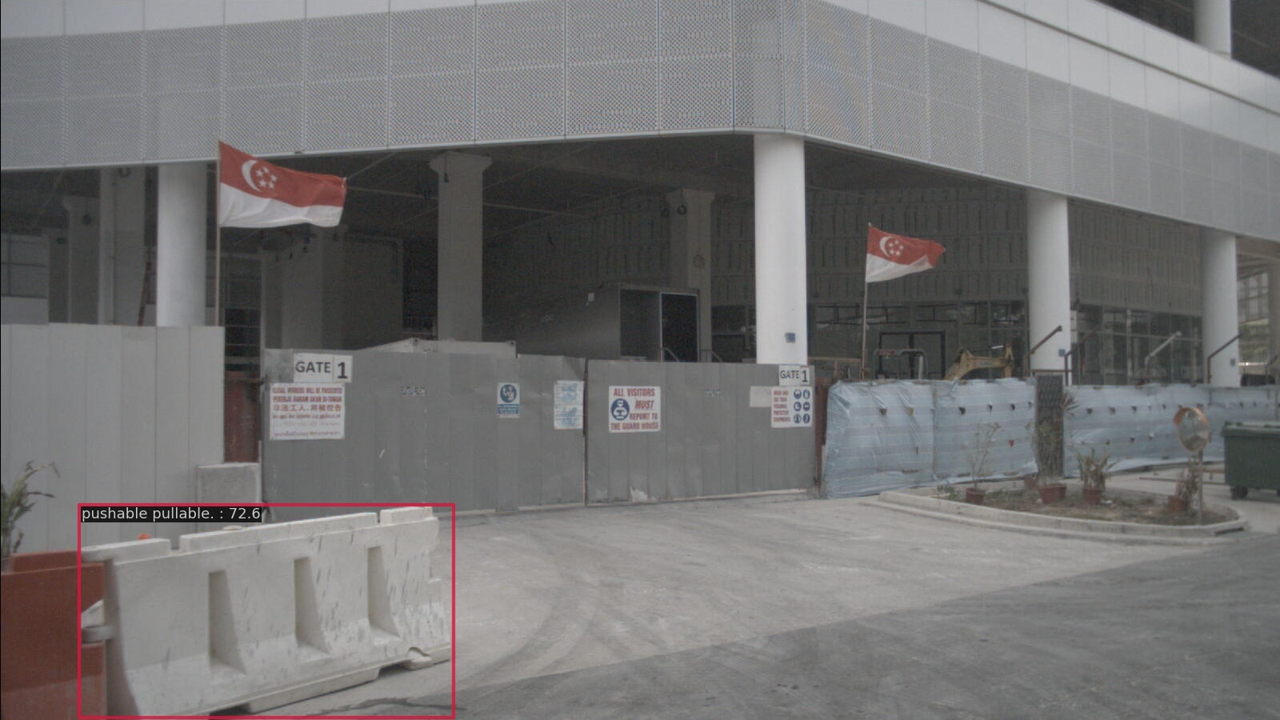}
        \includegraphics[width=0.49\textwidth]{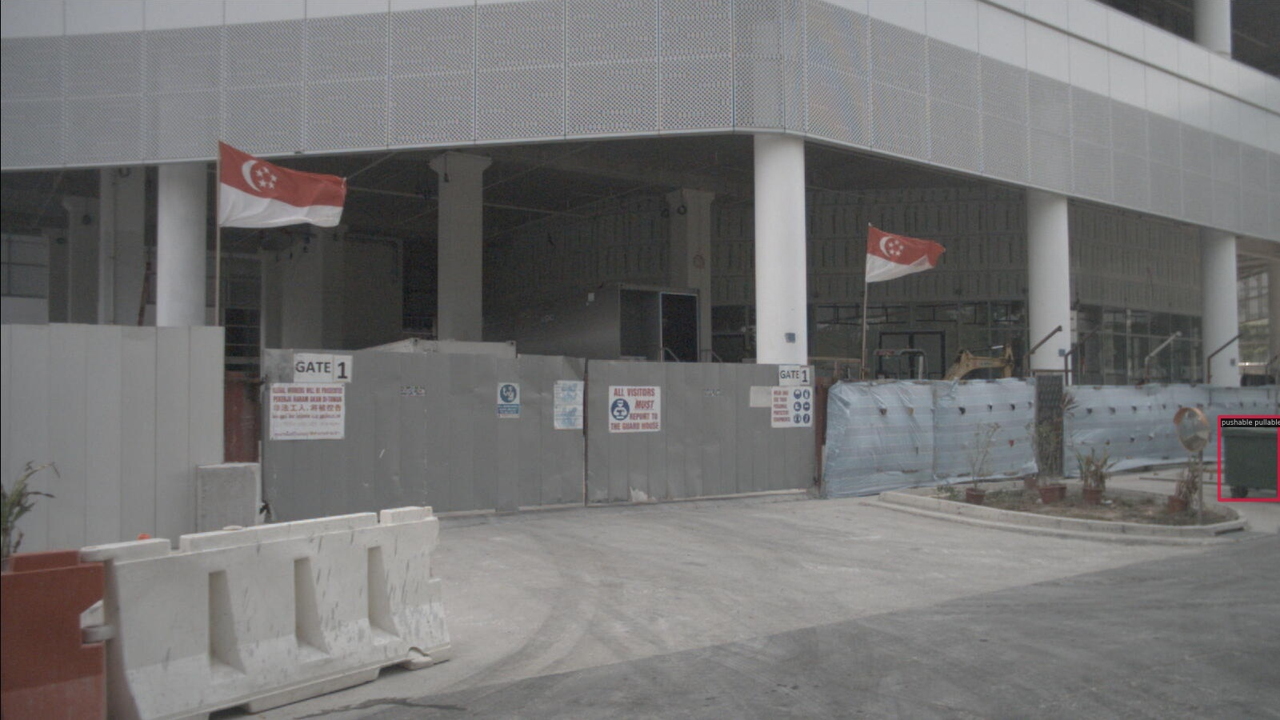}
        \caption{\textbf{left}: pushable pullable; \textbf{right}: pushable pullable garbage container (33139.jpg)}
    \end{subfigure}

    \begin{subfigure}[t]{1\linewidth}
        \centering
        \includegraphics[width=0.49\textwidth]{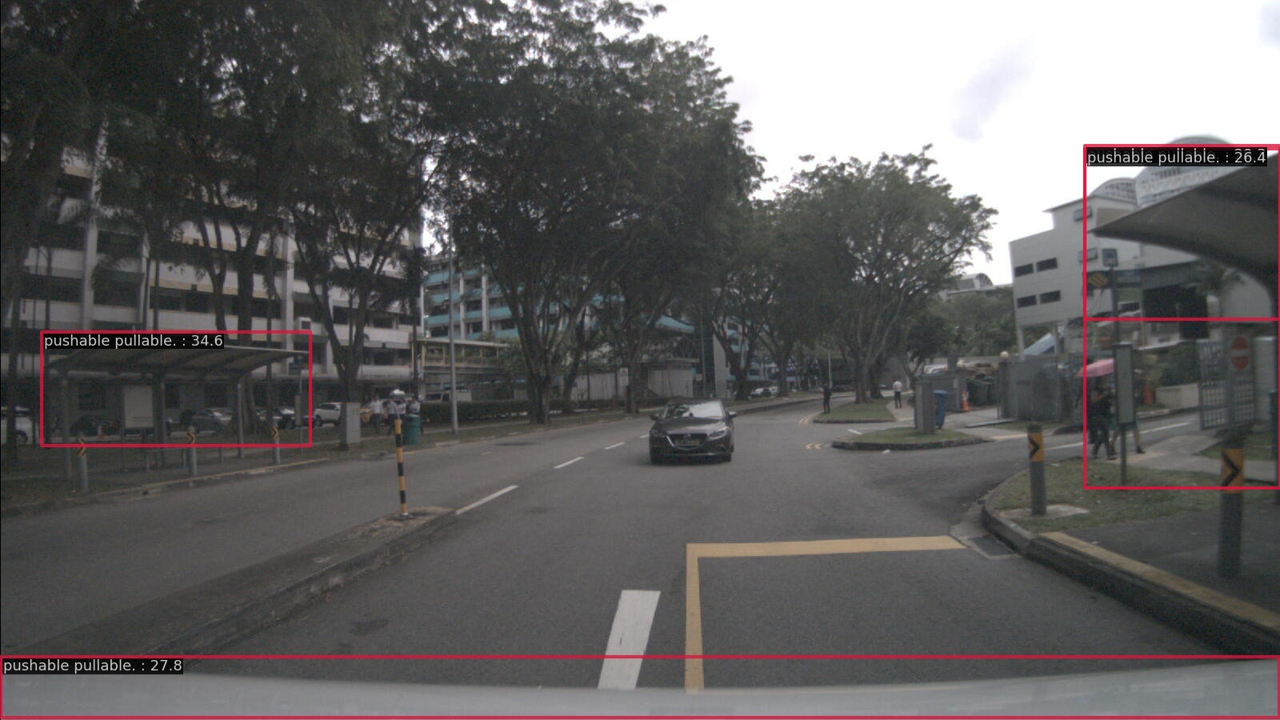}
        \includegraphics[width=0.49\textwidth]{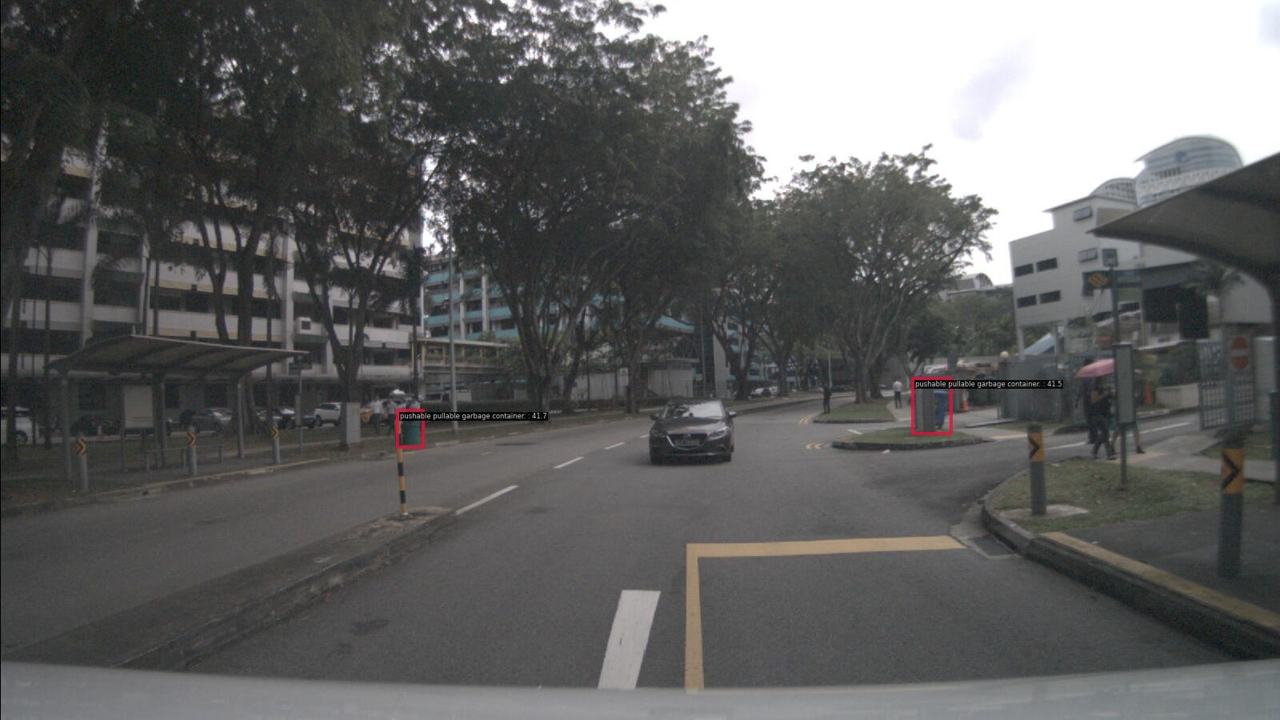}
        \caption{\textbf{left}: pushable pullable; \textbf{right}: pushable pullable garbage container (59104.jpg)}
    \end{subfigure}

    \begin{subfigure}[t]{1\linewidth}
        \centering
        \includegraphics[width=0.49\textwidth]{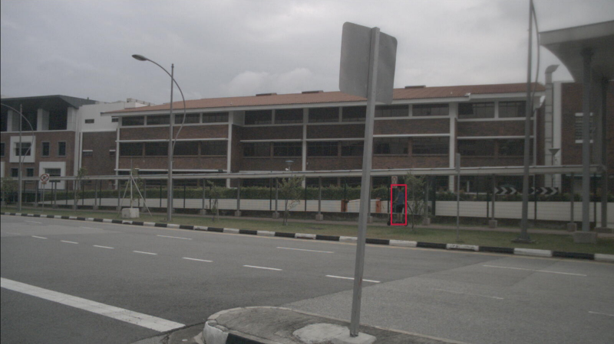}
        \includegraphics[width=0.49\textwidth]{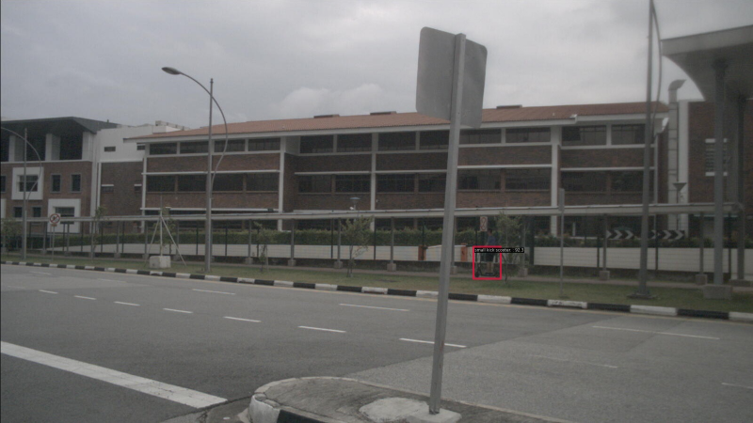}
        \caption{\textbf{left}: personal mobility; \textbf{right}: small kick scooter (2283.jpg)}
    \end{subfigure}

    \begin{subfigure}[t]{1\linewidth}
        \centering
        \includegraphics[width=0.49\textwidth]{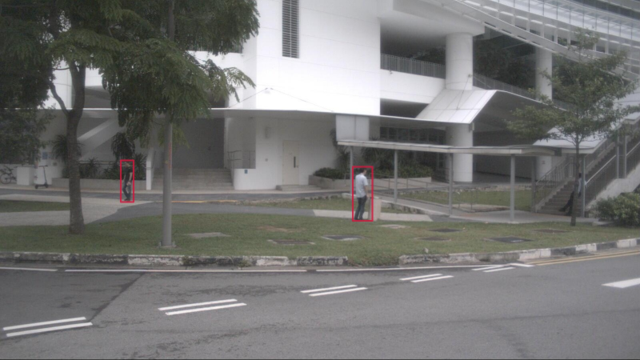}
        \includegraphics[width=0.49\textwidth]{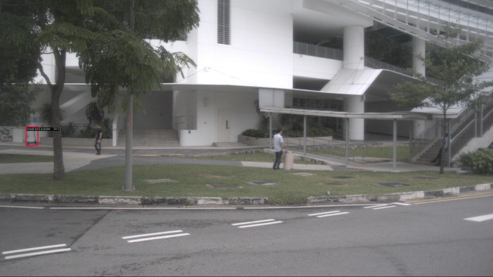}
        \caption{\textbf{left}: personal mobility; \textbf{right}: small kick scooter (5153.jpg)}
    \end{subfigure}

    \begin{subfigure}[t]{1\linewidth}
        \centering
        \includegraphics[width=0.49\textwidth]{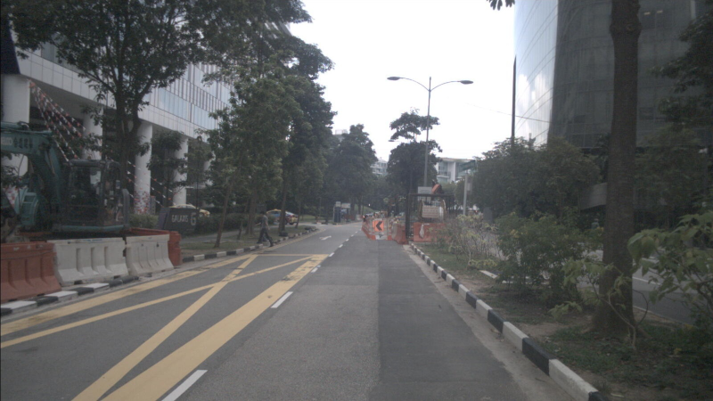}
        \includegraphics[width=0.49\textwidth]{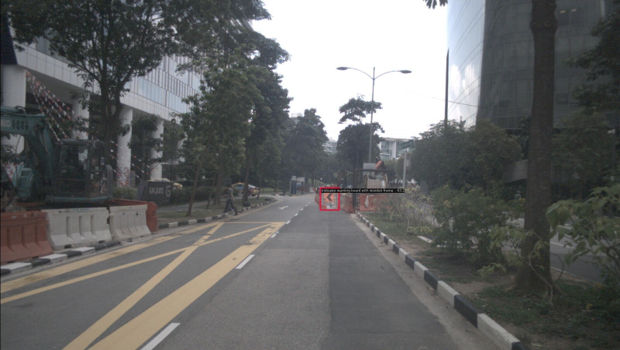}
        \caption{\textbf{left}: debris; \textbf{right}: indicator warning board with wooden frame (15429.jpg)}
    \end{subfigure}

    \begin{subfigure}[t]{1\linewidth}
        \centering
        \includegraphics[width=0.49\textwidth]{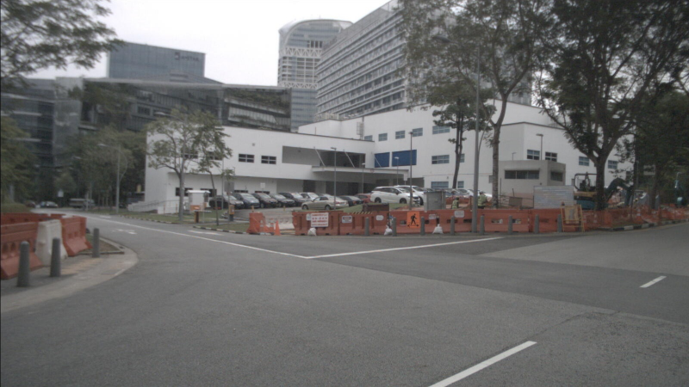}
        \includegraphics[width=0.49\textwidth]{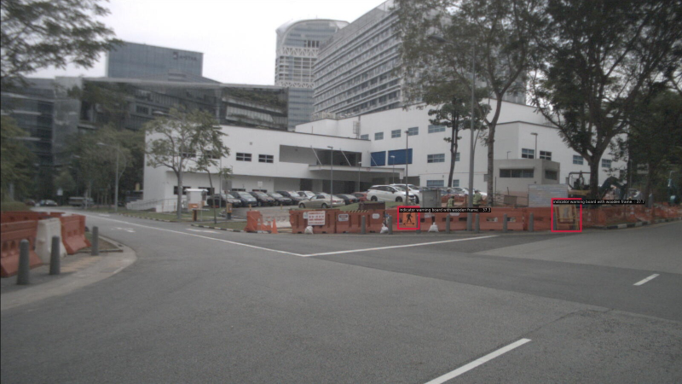}
        \caption{\textbf{left}: debris; \textbf{right}: indicator warning board with wooden frame (15937.jpg)}
    \end{subfigure}
    \caption{Visualizing examples: Given referential expressions about categories, VLMs can better detect new entities.}
\label{fig:case}
    
\end{figure}

\section{Experiment}
\textbf{Dataset.} The train and test datasets are provided by the organizers of the \href{https://eval.ai/web/challenges/challenge-page/2270/overview}{Foundational Few-Shot Object Detection Challenge.}

\textbf{Implementation Detail.}
We use ChatGPT-4 to generate 5 relevant prompt descriptions for each class. For the pseudo-label threshold \(\eta\), we set it to 0.3. The GLIP pre-training weights are selected from: 
\href{https://download.openmmlab.com/mmdetection/v3.0/glip/glip_atss_swin-l_fpn_dyhead_16xb2_ms-2x_funtune_coco/glip_atss_swin-l_fpn_dyhead_16xb2_ms-2x_funtune_coco_20230910_100800-e9be4274.pth}{mmdetection: GLIP-L}, which is pre-trained on the FourODs, GoldG, CC3M+12M, and SBU datasets.
The Grounding DINO pre-trained weights are selected from: 
\href{https://download.openmmlab.com/mmdetection/v3.0/mm_grounding_dino/grounding_dino_swin-l_pretrain_all/grounding_dino_swin-l_pretrain_all-56d69e78.pth}{mmdetection: MM-Grounding-DINO-L*}, which is pre-trained on the O365V2, OpenImageV6, GoldG, V3det, COCO2017, LVISV1, COCO2014, GRIT, RefCOCO, RefCOCO+, RefCOCOg, and gRefCOCO datasets.

\textbf{Result.} 
To validate the effectiveness of our approach, we apply VLM+ to both the pre-trained GLIP and Grounding DINO models. As depicted in Table~\ref{tab: compare}, the Grounding DINO model exhibits satisfactory performance in a zero-shot manner, owing to its extensive pre-trained on large datasets.
However, as illustrated in Table~\ref{tab:description}, VLM struggles to effectively understand certain class names when they are input as text. For example, the model's prediction accuracy for the term "debris" is 0. Conversely, utilizing modified referential expressions as input significantly enhances the prediction performance for this category. The incorporation of VLM+ leads to a notable improvement in performance, showcasing the effectiveness of our approach.

\textbf{Case Study.} 
We achieve improved performance by substituting category names with referential expressions as input text for VLMs. As illustrated in Figure~\ref{fig:case}, the original terms "personal mobility" and "pushable pullable" fail to accurately capture the semantic meaning of the objects, resulting in incorrect predictions by the VLM. Moreover, for the "debris" category, the model fails to generate any predictions, indicating a very low confidence level in this category. However, utilizing enhanced referential expressions for these categories as text prompts effectively mitigates the concept misalignment issues.

\section{Conclusion}
This report summarize our solution for the Foundational Few-Shot Object Detection Challenge (2024). 
By combining MM-LLM and VLMs, and utilizing a maximum IoU matching algorithm, we identify a referential expression aligned with the image concept for each category. Subsequently, we employ iterative pseudo-label generation and model optimization under these referential expressions. The final competition results demonstrate the effectiveness of our solution.

\bibliographystyle{ieee}
\bibliography{egpaper_final}

\end{document}